\documentclass[11pt]{article}
\usepackage{listings}
\usepackage[preprint]{acl}

\usepackage{times}
\usepackage{latexsym}
\usepackage{amsmath}
\usepackage{amsfonts}
\usepackage{tcolorbox}

\usepackage[T1]{fontenc}

\usepackage[utf8]{inputenc}

\usepackage{microtype}

\usepackage{inconsolata}
\usepackage{subcaption}

\usepackage{graphicx}
\usepackage{makecell}

\title{Stabilizing On-Policy Distillation for MLLM Reasoning with Global Normalization}

\author{
 \textbf{Dongze Hao},
 \textbf{Zhiwei Jin},
 \textbf{Chen Chen},
 \textbf{Haonan Lu}
 \\
 OPPO AI Center
 \\
 \small{
   \textbf{Correspondence:} {haodongze@oppo.com}
 }
}

\begin{document}
\maketitle
\begin{abstract}
On-policy distillation (OPD) has recently emerged as an important post-training paradigm. By using a stronger teacher model to provide dense, fine-grained supervision for sampled trajectories, OPD offers a clear advantage over reinforcement learning with verifiable rewards (RLVR), which typically depends on sparse binary or outcome-based environmental feedback. However, naive token-level distillation can suffer from gradient instability, due to magnitude misalignment in outlier states. To address this issue, we propose Globally Normalized Distillation Policy Optimization (GNDPO), a practical method that stabilizes optimization by transforming raw KL scores into batch-level relative advantages. This normalization effectively mitigates gradient explosions while retaining the benefits of token-level guidance. Experimental results show that GNDPO substantially improves training robustness and downstream performance across multimodal reasoning tasks. The code is released at \url{https://github.com/OPPO-Mente-Lab/GNDPO}.
\end{abstract}

\section{Introduction}
Post-training is a critical stage for injecting fine-grained reasoning abilities and domain-specific knowledge into large language models (LLMs) and multimodal large language models (MLLMs). A typical post-training pipeline consists of two stages: supervised fine-tuning (SFT) and reinforcement learning (RL) \cite{guo2025deepseek, zheng2025group, hu2025reinforce++, qin2025survey, song2026survey}. Although SFT provides direct linguistic supervision, it suffers from exposure bias at inference time. In contrast, on-policy RL methods alleviate distribution shift through online sampling, but they are often constrained by sequence-level sparse rewards. Such terminal-only feedback provides limited dense supervision for long-chain reasoning tasks, resulting in inefficient credit assignment and unstable training dynamics \cite{team2026kimi}.

Recently, on-policy distillation (OPD) has emerged as a promising alternative that bridges supervised fine-tuning and reinforcement learning by using the teacher model's logits as dense token-level supervision \cite{lu2025onpolicydistillation, deepseek2026v4, xiao2026mimo}. Despite its potential, existing OPD methods often suffer from substantial gradient instability. We observe that the raw KL divergence between the student and teacher models can fluctuate sharply when the student explores out-of-distribution (OOD) states. In particular, when the student generates tokens that the teacher assigns very low probability, the resulting log-ratio can induce large gradient spikes, disrupting the student's pretrained representations and leading to unstable or divergent training.

To address this issue, we introduce Globally Normalized Distillation Policy Optimization (GNDPO), a robust training framework that reformulates distillation as a relative optimization problem. Instead of directly optimizing raw teacher feedback, GNDPO converts token-level distillation rewards into relative advantages within each training batch. This global normalization mechanism dynamically rescales gradients and suppresses anomalous spikes, preventing optimization from being dominated by localized high-magnitude mismatches. As a result, GNDPO guides the student toward teacher-aligned distributions in a stable and controllable manner, providing a practical solution for large-scale model post-training.

To summarize, our contributions are threefold:

\begin{itemize}
\item We propose a globally normalized distillation mechanism that transforms raw teacher feedback into relative advantages, thereby stabilizing optimization in practical training settings.

\item We conduct extensive experiments across diverse multimodal reasoning benchmarks, showing that GNDPO consistently improves generation quality and training robustness over state-of-the-art RL and standard distillation baselines.
\end{itemize}


\section{Preliminary}
\subsection{On-policy RL}
Recent advancements of on-policy reinforcement learning methods such as GRPO \cite{guo2025deepseek} and GSPO \cite{zheng2025group}, has been increasingly favored in reinforcement learning for its efficiency and simplicity by replacing the traditional value model with a group-relative advantage estimation mechanism.

Formally, consider a given question $q$, where the behavior policy $\pi_{\theta_{\text{old}}}$ samples a group of $G$ responses $\{o_i\}_{i=1}^G$ and computes their rewards $\{r_i\}_{i=1}^G$.
The group-relative advantage for the $j$-th response, $A_j$, is subsequently derived by normalizing the aggregated rewards across the entire group:
\begin{equation}
A_i = \frac{r_i - \text{mean}\{r_1, \dots, r_G\}}{\text{std}\{r_1, \dots, r_G}
\end{equation}

Based on this advantage estimation, the GSPO optimization objective is formulated as:
\vspace{-0.5cm}
\begin{equation}
\begin{aligned}
\mathcal{J}_{\text{GSPO}}(\theta) = 
&\mathbb{E}_{(q, o_i) \sim D, \{o_i\}_{i=1}^G \sim \pi_{\theta_{\text{old}}}(\cdot|q)} \Biggl[ \frac{1}{G} \\
&\sum_{i=1}^G \frac{1}{|o_i|} \sum_{t=1}^{|o_i|} \min \Bigl( s_{i,t}(\theta) A_{i,t}, \\
&\text{clip}(s_{i,t}(\theta), 1 - \epsilon, 1 + \epsilon) A_{i,t} \Bigr) \Biggr]
\end{aligned}
\end{equation}
where the probability ratio is defined as $\text{sg}[s_i(\theta)] \cdot \frac{\pi_\theta(o_{i,t} \mid q, o_{i,<t})}{\text{sg}[\pi_\theta(o_{i,t} \mid q, o_{i,<t})]}$ and $\epsilon$ represents the clipping threshold. For simplicity, the KL-divergence regularization term is omitted in this formulation.

\subsection{On-Policy Distillation}
On-policy distillation generates trajectories with the student model and evaluates them at token-level granularity using a stronger teacher model. Unlike GRPO-style methods, which rely on sparse reward signals, on-policy distillation provides dense token-level supervision from the teacher model throughout the generated trajectory.

Formally, given a question $q$, the student policy $\pi_{\theta_{\text{old}}}$ first generates a response $o_i$ and computes the token-level logits. The teacher model then evaluates the same response and produces its corresponding logits. We use the reverse KL divergence between the student and teacher distributions as the reward signal for each token:
\begin{equation}
\hat{A}_{i,t} = -(\log \pi_\theta(o_i^{t} | q,o_j^{<t}) - \log \pi_{\text{teacher}}(o_i^{t} | q,o_j^{<t}))
\end{equation}

We define the token-level reward as the negative reverse KL divergence between the student and teacher distributions, encouraging the student policy to match the teacher. This reward reaches its maximum when the student distribution exactly recovers the teacher distribution. The resulting rewards are then used in an importance-sampling-based RL objective to update the student model.

\begin{figure}[t]
  \centering
  \begin{subfigure}[t]{\columnwidth}
    \centering
    \includegraphics[width=\linewidth]{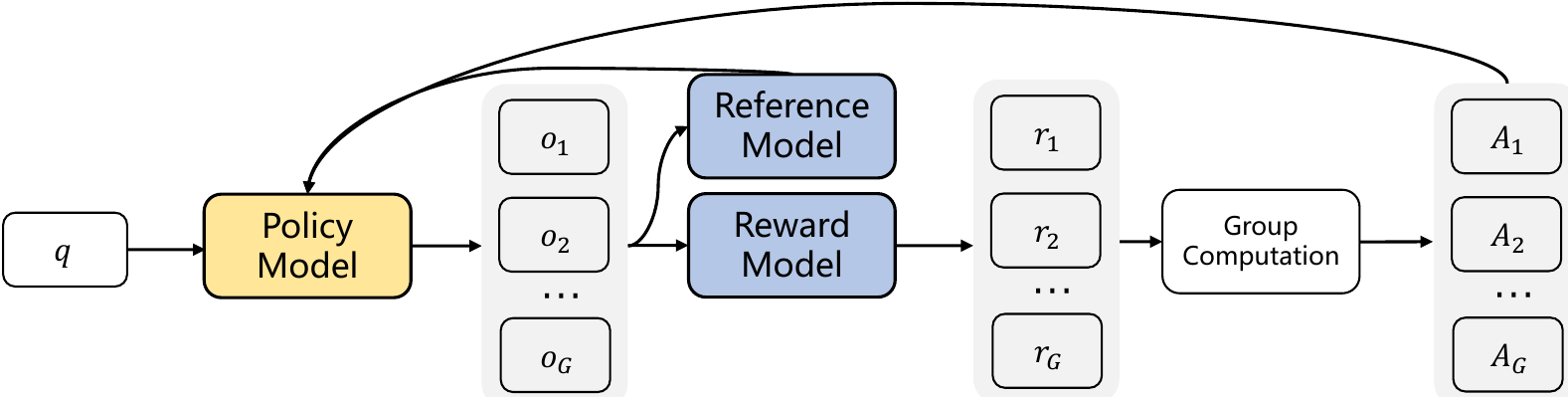}
    \caption{GSPO}
    \label{fig:method:grpo}
  \end{subfigure}

  \vspace{0.5em}

  \begin{subfigure}[t]{\columnwidth}
    \centering
    \includegraphics[width=\linewidth]{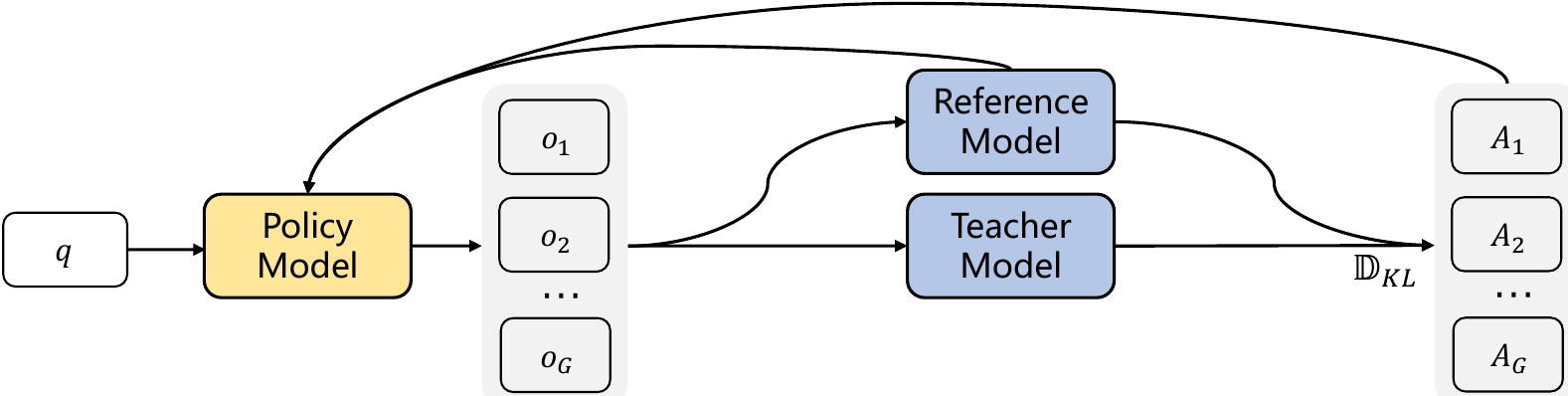}
    \caption{OPD}
    \label{fig:method:opd}
  \end{subfigure}

  \vspace{0.5em}

  \begin{subfigure}[t]{\columnwidth}
    \centering
    \includegraphics[width=\linewidth]{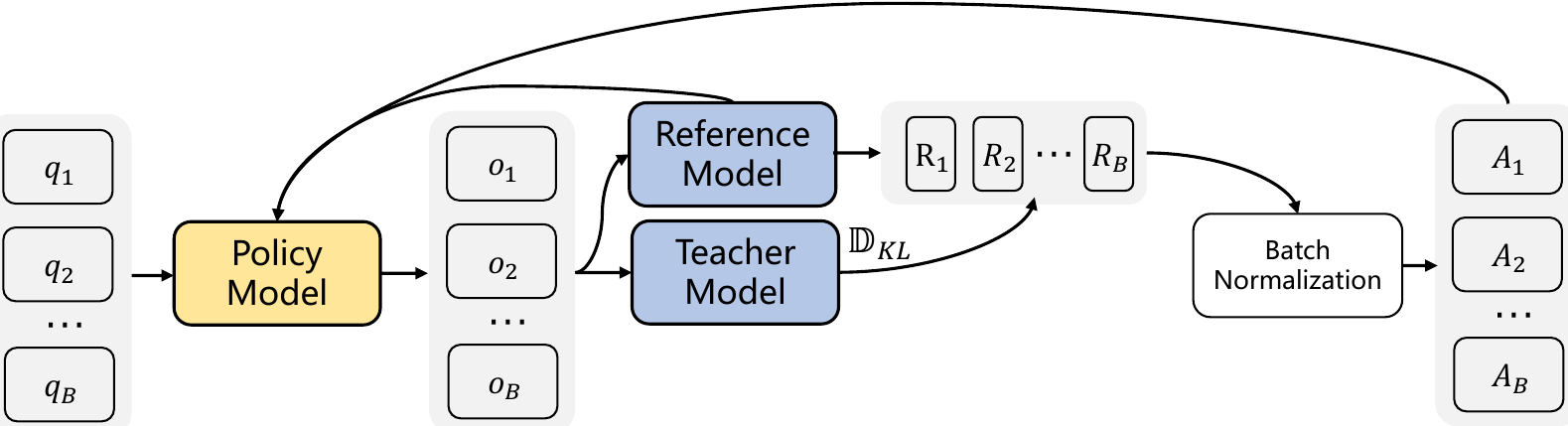}
    \caption{GNDPO}
    \label{fig:method:gndpo}
  \end{subfigure}

  \caption{The comparison of GSPO, OPD and our proposed GNDPO method. GSPO relies on sparse, sequence-level rewards, which makes detailed credit assignment difficult. OPD provides dense, token-level guidance using raw KL divergence, but it is prone to gradient explosion. Our GNDPO solves this problem by normalizing the rewards within each batch, ensuring both high signal density and stable training dynamics. }
  \vspace{-0.5cm}
  \label{fig:method}
\end{figure}

\section{Method}
\subsection{Motivation}
Although token-level dense rewards provide richer supervision than sequence-level sparse rewards, such as those used in GSPO, they can introduce substantial gradient instability during on-policy exploration.

In particular, when the student policy visits out-of-distribution (OOD) contexts or generates low-frequency tokens to which the teacher assigns near-zero probability,
$\pi_{\text{teacher}}(o_j^{t} \mid q, o_j^{<t}) \to 0$,
the log-ratio term
$\log \frac{\pi_\theta(o_j^{t} \mid q, o_j^{<t})}{\pi_{\text{teacher}}(o_j^{t} \mid q, o_j^{<t})}$
can become extremely large. As a result, the raw distillation reward $\hat{A}_{i,t}$ may decrease sharply, producing disproportionately large gradient updates for a small number of tokens:$
\nabla_\theta \mathcal{L}(\theta) \propto
\nabla_\theta \log \pi_\theta(o_j^{t} \mid q, o_j^{<t}) \cdot \hat{A}_{i,t}
$
Directly optimizing these raw KL-based scores can cause the policy gradient to be dominated by a small subset of high-entropy or highly mismatched tokens. This magnitude misalignment across contexts disrupts stable policy updates and can lead to training divergence.

\subsection{Globally Normalized Distillation Policy Optimization}
To mitigate the gradient explosion caused by localized logit discrepancies, we propose Globally Normalized Distillation Policy Optimization (GNDPO). Instead of assigning the raw KL divergence directly as the optimization landscape, we calibrate the token-level rewards into a relative advantage within each training batch $\mathcal{B}$.

For each token $o_{i,t}$ in the batch, its normalized distillation advantage $\hat{A}_{i,t}^\text{norm}$ is formulated as:
\vspace{-0.5cm}
\begin{equation}
\hat{A}_{i,t}^\text{norm} = \frac{\hat{A}_{i,t}-\text{mean}(A|A\in \mathcal{B})}{\text{std}(A|A\in \mathcal{B})+ \epsilon} 
\end{equation}

This global normalization mechanism offers two critical benefits:Adaptive Gradient Rescaling: 

\begin{enumerate}
    \item When catastrophic logit mismatches occur, the extreme negative rewards expand the batch standard deviation $\sigma_{\mathcal{B}}$. This automatically scales down the effective gradient step size, preventing anomalous tokens from rupturing the shared representations.
    \item Context-Agnostic Scale Calibration: By shifting the baseline via $\mu_{\mathcal{B}}$, the objective transitions from tracking absolute distribution matching to optimizing relative token quality. This balances the optimization contribution across heterogeneous contexts of varying generation difficulties.
\end{enumerate}











\section{Experiments}
\subsection{Experimental Setup}
\paragraph{Models and Datasets.} We conduct experiments with the InternVL3.5 model family \cite{wang2025internvl3} at three scales: InternVL3.5-1B-Instruct, InternVL3.5-2B-Instruct, and InternVL3.5-4B-Instruct. We use InternVL3.5-8B as the teacher model for OPD and GNDPO. For training data, we use the multimodal mathematical reasoning dataset Geometry3K \cite{lu2021inter}. We evaluate the model on various challenging  multimodal reasoning benchmarks: MathVista \cite{lu2024mathvista}, MMMU \cite{yue2024mmmu}, MathVision \cite{wang2024measuring}, MathVerse \cite{zhang2024mathverse}, DynaMath \cite{zou2025dynamath}, Wemath \cite{qiao2025we}, LogicVista \cite{xiao2024logicvista}.

\paragraph{Baselines.} We compare our method with following baselines: (1) GSPO, group sequence relative policy optimization with binary outcome rewards verified against
ground-truth answers. (2) GSPO-OPD, group sequence relative policy optimization with on policy distill rewards by the teacher model.

\paragraph{Implementation details.} We implement our method based on the VERL \cite{sheng2025hybridflow} framework. For GSPO, we sample 16 responses per problem. During training, the model’s maximum context length is set to 24576, with a maximum prompt length of
8192 and a maximum response length of 16384. During evaluation, the model’s maximum context length is set to 32768. We use Adam optimizer with a constant learning rate
of 1e-6. The training batch size is set to 32, and we sample 16 responses per problem with a sampling temperature of 1.0. All experiments are conducted on 8 NVIDIA H20 100GB GPUs. More experimental details are in Appendix \ref{sec:appendix}.

\begin{table*}
  \centering
  \resizebox{\linewidth}{!}{
  \begin{tabular}{lcccccccc}
    \hline
     &\makecell{MathVista\\MINI}	& \makecell{MMMU\\VAL}	&MathVision	& \makecell{MathVerse\\(vision-only)}	& \makecell{DynaMath\\
(worst case)} &WeMath	&LogicVista	 &Overall  \\
    \hline
    \textit{InternVL3.5-8B} & 78.4&73.4&56.8&61.5&37.7&57.0&57.3&60.3\\
    \hline
    \textit{InternVL3.5-4B}\\
    Base (Instruct)&71.4&64.3&40.5&50.0&30.7&35.6&53.5&49.4\\
    +GSPO&71.4	&67.2	&48.7	&52.7	&31.1	&42.6 &53.5&52.5\\
    +OPD&\textbf{73.3}	&67.2	&50.5	&52.9	&\textbf{32.7}&	47.2&	53.9& 53.9\\
    +GNDPO&73.0	&\textbf{68.3}	&\textbf{51.6}	&\textbf{53.2}	&32.5   &\textbf{48.8}	&\textbf{54.4} &\textbf{54.5}\\
    \hline
    \textit{InternVL3.5-2B}\\
    Base (Instruct)&60.8&53.0&27.0&39.6&19.8&28.1&41.2&38.5\\
    +GSPO&62.3	&54.7	&35.9	&43.2	&19.8	&28.2	&39.6&40.5\\
    +OPD&64.7	&57.4	&37.9	&42.0	&\textbf{24.4}	&\textbf{34.9}	&39.6&42.9\\
    +GNDPO&\textbf{66.0}	&\textbf{57.7}	&\textbf{38.4}	&\textbf{43.7}	&23.8	&34.8	&\textbf{41.6} & \textbf{43.7}\\
    \hline
    \textit{InternVL3.5-1B}\\
    Base (Instruct)&37.2 &48.6 &15.8 &27.0 &8.4 &13.9 &29.1 &25.7\\
    +GSPO&52.2	&39.9	&21.6	&28.2	&8.4	&14.0	&26.6& 27.3\\
    +OPD&55.9	&47.1	&24.0	&28.3	&11.8	&23.1	&30.9&31.6\\
    +GNDPO&\textbf{56.6}	&\textbf{47.6}	&\textbf{24.5}	&\textbf{28.4}	&\textbf{12.0}	&\textbf{23.7}	&\textbf{31.3}&\textbf{32.0}\\
    \hline
  \end{tabular}}
  \caption{Multimodal reasoning performance across different models with various optimization methods. \textbf{Bold} indicates the best result. "Overall" represents the average score across the seven selected benchmarks.}
  \label{tab:mainresults}
\end{table*}

\begin{figure*}[t]
  \centering
  \begin{subfigure}[t]{0.65\columnwidth}
    \centering
    \includegraphics[width=\linewidth]{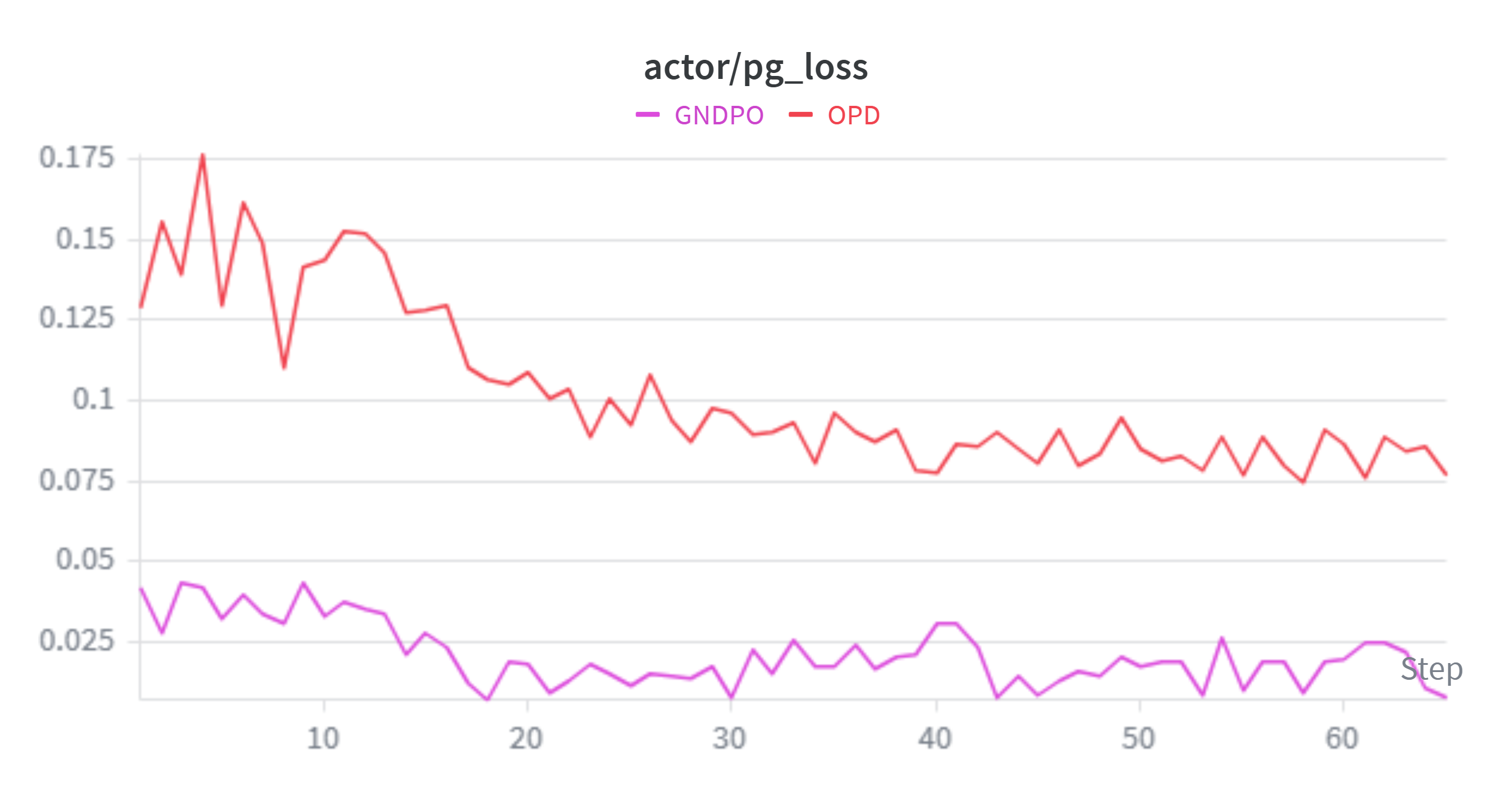}
    \caption{Training loss dynamics comparison.}
  \end{subfigure}
  \hfill
  \begin{subfigure}[t]{0.65\columnwidth}
    \centering
    \includegraphics[width=\linewidth]{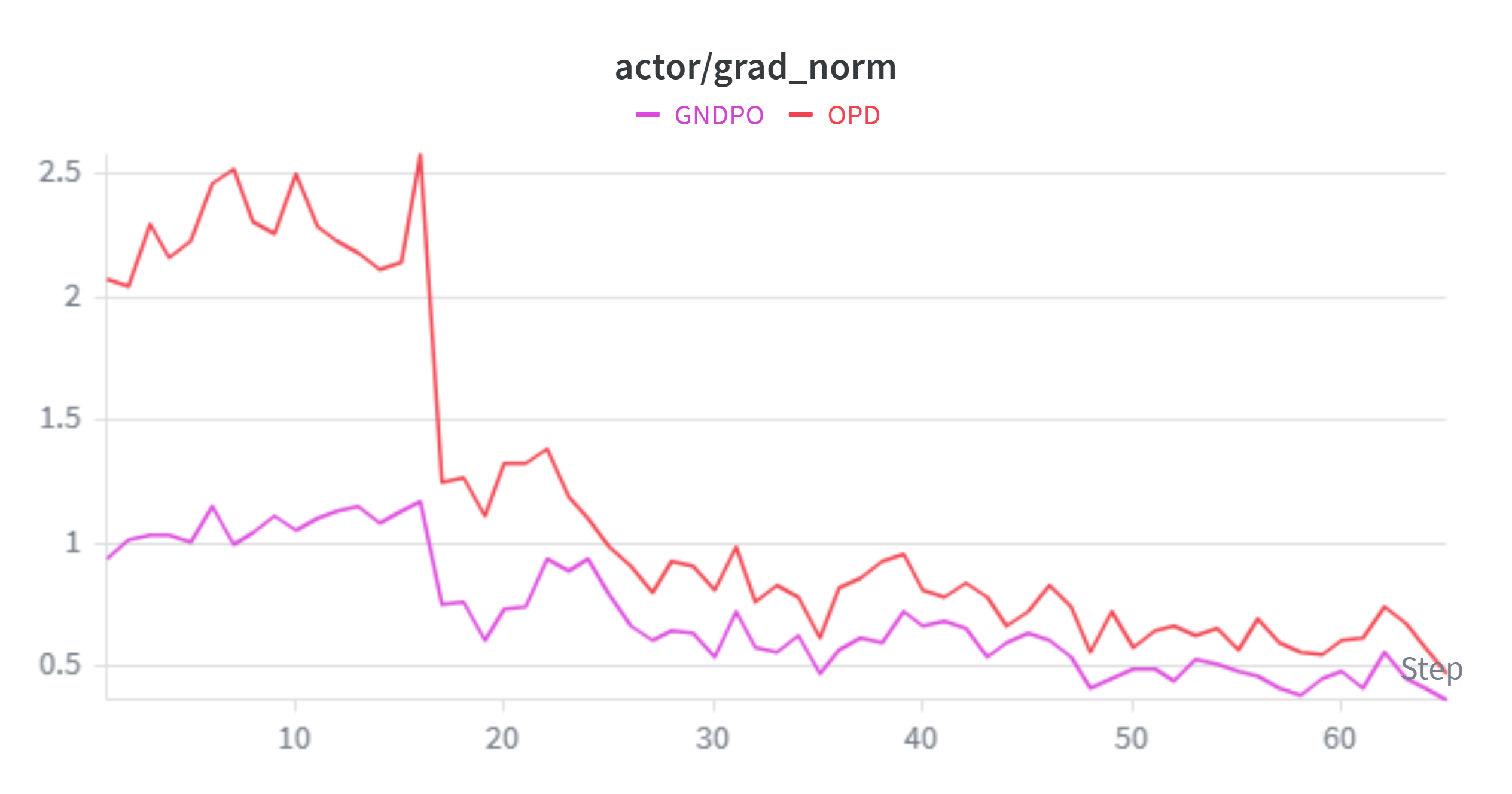}
    \caption{Training grad\_norm dynamics comparison.}
    \label{fig:method:opd}
  \end{subfigure}
  \hfill
  \begin{subfigure}[t]{0.65\columnwidth}
    \centering
    \includegraphics[width=\linewidth]{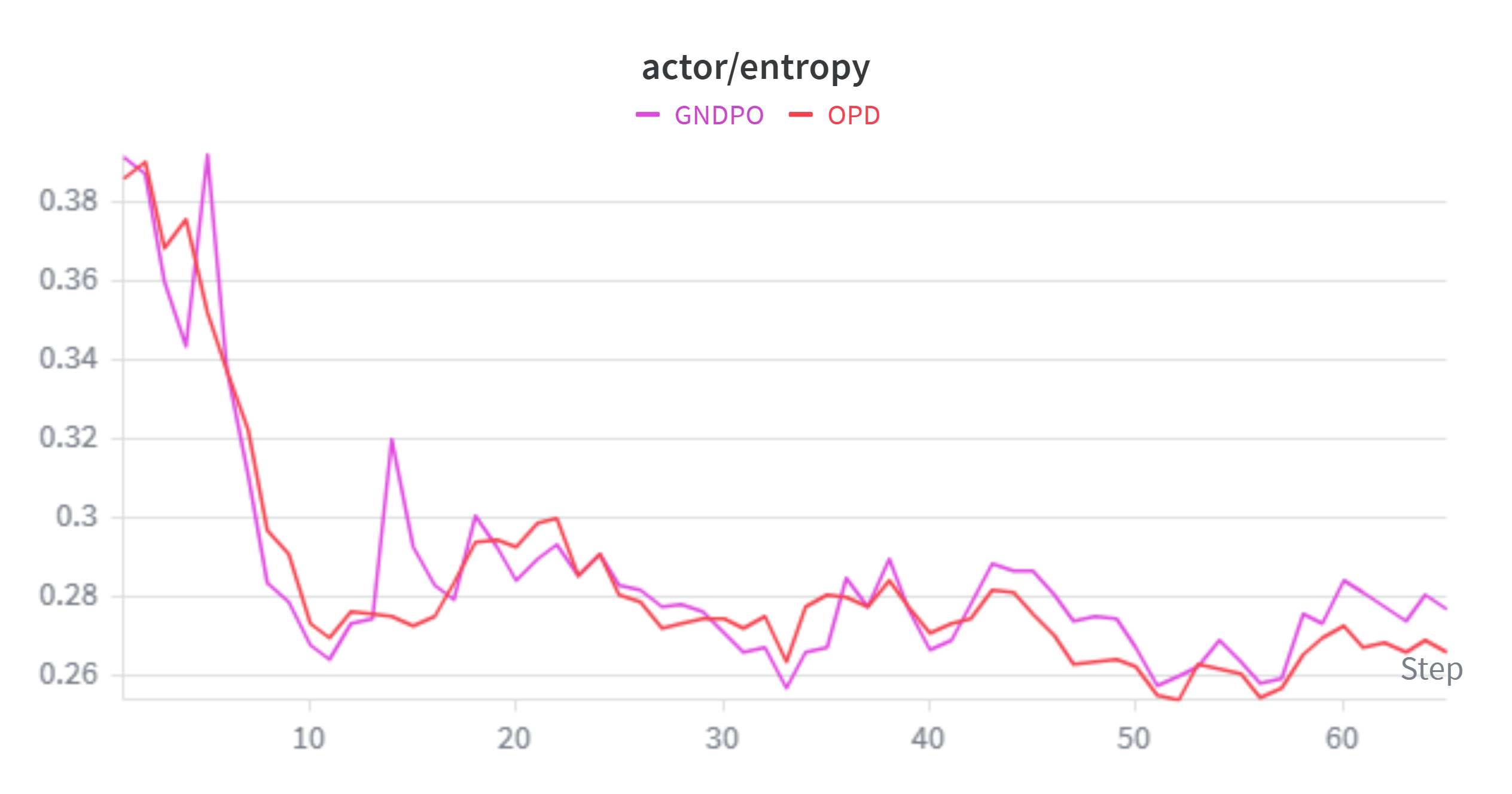}
    \caption{Training entropy dynamics comparison.}
    \label{fig:method:gndpo}
  \end{subfigure}
  \caption{Training dynamics of OPD method and GNDPO method.}
  \vspace{-0.5cm}
  \label{dynamics}
\end{figure*}

\subsection{Main Results}
Table~\ref{tab:mainresults} reports the evaluation results across seven multimodal reasoning benchmarks. GNDPO consistently outperforms the baseline methods in average accuracy across scales. Specifically, for InternVL3.5-4B, GNDPO leverages stabilized token-level advantages to achieve notable gains on complex reasoning datasets, including MMMU (+1.1\%), MathVision (+1.1\%), and WeMath (+1.6\%), where precise reasoning chains are crucial. For InternVL3.5-2B, our method substantially boosts performance on MathVista (+1.3\%), MathVision (+0.5\%), and MathVerse (+1.7\%). Concurrently, for the parameter-constrained InternVL3.5-1B, GNDPO drives distinct improvements on MathVista (+0.7\%) and MMMU (+0.5\%), proving that robust reward calibration is vital even for smaller models.
While individual performance gains vary due to the baseline models' differing domain capacities, the consistent aggregate improvements underscore GNDPO's effectiveness in achieving stable behavioral alignment.

\subsection{Training Dynamics}
Figure \ref{dynamics} illustrates the training dynamics over 65 optimization steps. As shown in Figure \ref{dynamics}(a), Standard OPD exhibits severe volatility due to raw, uncalibrated reward fluctuations, whereas GNDPO maintains a significantly lower and smoother loss profile throughout training. Figure \ref{dynamics}(b) reveals that standard OPD suffers from a catastrophic gradient spike ($\ell_2\text{ norm} > 2.5$) around step 15 caused by localized logit mismatches, while GNDPO successfully suppresses these anomalous spikes via global batch normalization. Figure \ref{dynamics}(c) shows that GNDPO stabilizes the gradient scale without sacrificing the model's exploratory capacity.


\section{Conclusion}
In this paper, we introduced Globally Normalized Distillation Policy Optimization (GNDPO) to address the issue of gradient instability in on-policy token-level distillation. By reformulating raw, context-dependent teacher feedback into relative batch-level advantages, GNDPO successfully suppresses catastrophic gradient spikes caused by localized logit discrepancies. Extensive experiments across multiple multimodal reasoning benchmarks demonstrate that our framework consistently outperforms standard OPD and RLVR baselines in both average accuracy and training robustness.

\section*{Limitations}
Despite its empirical success and enhanced stability, GNDPO exhibits a few limitations: first, since our global normalization relies on empirical statistics calculated over a training mini-batch, its stabilization capability may slightly degrade under extremely small batch sizes due to increased statistical variance; second, as an on-policy paradigm, it inherits the standard computational overhead of continuous token generation during rollouts, meaning the training wall-clock time remains bounded by model inference speeds; finally, while we successfully verified its efficacy on core mathematical and multimodal reasoning tasks, GNDPO's behavioral dynamics in open-ended creative text generation or multi-turn conversational alignment remain to be extensively explored.


\bibliography{custom}

\appendix

\section{Appendix}
\label{sec:appendix}

\subsection{Experimental Details}
We provide the training and evaluation configurations in Table \ref{tab:train_pram} and Table \ref{tab:test_pram}. Both GSPO, OPD and GNDPO methods use the same base hyperparameters where applicable to ensure fair comparison. We use the vlmevalkit \cite{duan2024vlmevalkit} framework to evaluate the model performance. 

\begin{table}
  \centering
  \begin{tabular}{lc}
    \hline
    \textbf{Parameter} & \textbf{Value} \\
    \hline
    Learning rate     & $1 \times 10^{-6}$          \\
    Train batch size     & 32           \\
    PPO mini-batch size     & 32           \\
    Number of rollouts     & 16          \\
    Sampling temperature     & 1.0          \\
    Max prompt length      & 8192            \\
    Max response length     & 16384          \\
    Training steps     & 65          \\
    KL coefficient $\beta$ & - \\
    \hline
  \end{tabular}
  \caption{Training hyper-parameters.}
  \label{tab:train_pram}
\end{table}

\begin{table}
  \centering
  \begin{tabular}{lc}
    \hline
    \textbf{Parameter} & \textbf{Value} \\
    \hline
    Max new tokens     & 32768          \\
    Temperature     & 0.6         \\
    Top-p     & 0.95          \\
    Top-k     & -1          \\
    \hline
  \end{tabular}
  \caption{Evaluation hyper-parameters.}
  \label{tab:test_pram}
\end{table}

\subsection{Prompt}
The reasoning prompt used in the model is shown as follows:
\begin{tcolorbox}[colback=gray!5, colframe=black!45,
    fonttitle=\bfseries, title= Prompt used for Answering Questions]
\small

\begin{lstlisting}[breaklines=true,breakindent=0pt,columns=fullflexible]
You FIRST think about the reasoning process as an internal monologue and then provide the final answer. The reasoning process MUST BE enclosed within <think> </think> tags. The final answer MUST BE put in \\boxed{}
\end{lstlisting}
\end{tcolorbox}

\end{document}